\documentclass[conference]{IEEEtran}
\IEEEoverridecommandlockouts
\usepackage{cite}
\usepackage{amsmath,amssymb,amsfonts}
\usepackage{algorithmic}
\usepackage{graphicx}
\usepackage{textcomp}
\usepackage{xcolor}
\usepackage{colortbl}
\usepackage{multirow}
\usepackage{threeparttable}
\def\BibTeX{{\rm B\kern-.05em{\sc i\kern-.025em b}\kern-.08em
    T\kern-.1667em\lower.7ex\hbox{E}\kern-.125emX}}
\begin{document}

\title{A Centralized-Distributed Transfer Model for Cross-Domain Recommendation 
Based on Multi-Source Heterogeneous Transfer Learning\\
}

\author{\IEEEauthorblockN{Ke Xu, Ziliang Wang, Wei Zheng*, Yuhao Ma, Chenglin Wang,
Nengxue Jiang, Cai Cao}
\thanks{*Wei Zheng is the corresponding author.}
\IEEEauthorblockA{\textit{Hangzhou NetEase Cloud Music Technology Co., Ltd.} \\
\textit{Phase II, Netease building, No. 599, Wangshang Road, 
Hangzhou Shi, Zhejiang Province, China} \\
Email:\{xuke04,wangziliang,zhengwei03,mayuhao1,wangchenglin,jiangnengxue,caocai\}@corp.netease.com}
}

\maketitle

\thispagestyle{plain}
\pagestyle{plain}

\begin{abstract}
    Cross-domain recommendation (CDR) methods are proposed to tackle the sparsity problem 
    in click through rate (CTR) estimation. Existing CDR methods directly transfer 
    knowledge from the source domains to the target domain and ignore the heterogeneities 
    among domains, including feature dimensional heterogeneity and latent space 
    heterogeneity, which may lead to negative transfer. Besides, most of the existing methods 
    are based on single-source transfer, which cannot simultaneously utilize knowledge 
    from multiple source domains to further improve the model performance in the target 
    domain. In this paper, we propose a centralized-distributed transfer model (CDTM) 
    for CDR based on multi-source heterogeneous transfer learning. To address the issue of feature 
    dimension heterogeneity, we build a dual embedding structure: domain specific 
    embedding (DSE) and global shared embedding (GSE) to model the 
    feature representation in the single domain and the commonalities in the global space,
    separately. To solve the latent space heterogeneity, the transfer matrix and attention mechanism 
    are used to map and combine DSE and GSE adaptively. Extensive offline and online experiments 
    demonstrate the effectiveness of our model.
\end{abstract}


\section{Introduction\label{Introduction}}

Traditional click through rate (CTR) models\cite{guo2017deepfm,zhou2018deep} 
mainly focus on the recommendations for a single scenario or a single domain. 
A well-trained model requires sufficient data of the current domain. 
However, in the recommendation system and online advertising system, 
there are many domains to be served, some of which may suffer from data sparsity 
problem even in the top advertising platforms. This data sparsity problem raises 
a series of challenges to the performance of traditional CTR model. 
To solve this problem, the idea of transfer learning is introduced. The CTR method that 
integrates the idea of transfer learning is called cross-domain recommendation (CDR). 
In recent years, many CDR methods are proposed\cite{zang2021survey}. 
However, these existing methods still have two obvious drawbacks.

Firstly, most of these methods are based on single source transfer. In other 
words, for the target domain, the knowledge from only one source domain 
will be transferred. However, in the advertising system, domains with different 
ad types (contract ads or bid ads) or different ad display areas (called ``flight" 
% NetEase Cloud Music
in this paper) can transfer different knowledge to the target domain. 
This knowledge can improve the performance of the CTR model in 
the target domain. Therefore, how to simultaneously transfer multi-source 
information to the target domain and improve the recommendation accuracy  
on the target domain is a challenge.

% \begin{figure*}[htbp]
%     \centerline{\includegraphics[width=0.8\linewidth]{img-new.png}}
%     \caption{An example of multi-source transfer in Anonymous advertising platform.}
%     \label{fig1}
% \end{figure*}
% NetEase Cloud Music
Secondly, existing CDR methods assume that all the domains have the same features and 
the heterogeneities between source domain and target domain are ignored, including 
feature dimensional heterogeneity and latent space heterogeneity. Different domains 
may have different feature dimensions, which is called dimension 
heterogeneity in this paper. Actually, each domain includes two types of features: 
one type is the transferable features (e.g. age, gender), which can be shared with some other domains, 
and the other type is the nontransferable features, which cannot be shared and are 
unique to the domain (e.g. users' clicked ads in this domain). 
And the numbers of transferable features or the nontransferable 
features in different domains may be different. Moreover, even the same features may 
have different distributions in different domains, which is called latent space 
heterogeneity. Networks of these domains' models cannot directly be shared with each 
other, which is widely used by existing CDR methods. These two heterogeneities 
limit the application of existing methods and even lead to negative 
transfer, which may weaken the performance of the target domain model.

For the above two problems, we proposes a centralized-distributed transfer 
model (CDTM) for CDR based on multi-source heterogeneous transfer learning. 
The main contributions of our work are summarized as follows:

1. A centralized-distributed transfer model for CDR is proposed. The proposed model can be also 
extended to scenarios with more domains and simultaneously improve the 
performance of multiple domain models.

2. The proposed model constructs a dual embedding structure: domain specific embedding (DSE)
and global shared embedding (GSE) are used to model the unique feature representation of 
single domain and the global feature representation of all domains, respectively. 
The combination attention is developed to adaptively combine dual embedding of transferable features.

3. The proposed model utilizes the transfer matrix to map the GSE into a shared latent space 
with DSE to deal with the heterogeneous problem in cross-domain recommendation. 
And an auxiliary loss is constructed to help the optimization of the transfer matrix.

% 4. The proposed model can be also extended to scenarios with more domains and 
% simultaneously improve the performance of multiple domain models.

% NetEase Cloud Music

4. Extensive offline and online experiments are conducted based on real-world commercial 
data, which demonstrates the effectiveness and robustness of the proposed model. 

% This paper is organized as follows: The related work is introduced in Section~\ref{Related Work}. 
% The details of the proposed CDTM is presented in Section~\ref{Model Description}. The results and analysis 
% of offline and online experiments are shown in Section~\ref{Experiments} and Section~\ref{Online A/B Test} respectively. 
% Finally, we conclude the paper in Section~\ref{Conclusions}.

\section{Related Work\label{Related Work}}

\subsection{CTR Estimation}

CTR estimation refers to estimating the click probability 
for a given exposure and it plays an important role in the advertising system. 
The linear LR\cite{richardson2007predicting} model is first applied to CTR estimation. But the LR model 
lacks the ability to learn feature interactions. To address this problem, 
FM\cite{rendle2010factorization} model is proposed to learn the second-order feature interactions, 
and FFM\cite{juan2016field} further improved this idea. After that, deep neural networks methods 
\cite{guo2017deepfm,qu2016product,cheng2016wide,wang2017deep,cheng2020adaptive,wang2021dcn} 
are applied to CTR prediction to automatically learn high-order nonlinear feature interactions. 
In addition, some models are proposed to model the user interests, such as DIN\cite{zhou2018deep},
DIEN\cite{zhou2019deep} and MIND\cite{li2019multi}. MIMN\cite{pi2019practice} and SIM\cite{pi2020search} 
further develops this idea to model long-term user interest.

The previous work achieves good performance in the single-domain. 
However, well-trained models require sufficient data, which is not 
available in some domains with sparse data. Therefore, CDR 
methods are proposed to solve this problem.

\subsection{Cross Domain Recommendation\label{Cross Domain Recommendatio}}

In recent years, many CDR methods are proposed
to transfer knowledge from the source domain to the target domain for providing 
enhanced recommendations. Reference\cite{man2017cross} first proposes an embedding and mapping 
framework for CDR and other work\cite{zhu2019dtcdr,qiu2019cross,zhu2020deep,zhu2021unified,yuan2019darec} 
develop this idea. Some work is devoted to achieving bidirectional knowledge transfer 
between two domains, such as CoNet\cite{hu2018conet}, DTCDR\cite{zhu2019dtcdr}, 
DDTCDR\cite{li2020ddtcdr}, CAN\cite{qiu2019cross} and GA-DTCDR\cite{zhu2021unified}. 
Other methods\cite{ouyang2020minet,li2021dual,zhu2022personalized} focus on user behavior interest in 
CDR by transferring sequence features of domains. Recently, methods for multi-domain 
recommendation (MDR) are proposed\cite{he2020dadnn,sheng2021one}, 
these methods tackle multiple domians CTR estimation using one model and 
network-sharing is the main characteristics of these methods. 

% There is a slight difference between MDR and CDR. The goal of MDR is to simultaneously 
% improve the performance of multiple domains, which often have enough data to train models individually. 
% While the core issue of CDR is to utilize domains with sufficient data to improve the 
% performance of target domains, which suffer from the problem of data sparsity. 
% From this point of view, MDR is a special case of CDR. 
% In this paper, we focus on the CDR problem.

Despite the great success made by these methods, there are also some problems to be solved. 
Firstly, they mostly transfer knowledge from one source and multi-source transfer 
is seldom considered. Secondly, the existing CDR methods assume that all the domains 
have the same features, and the information transfer is based on sharing networks. 
Actually, there are feature dimension heterogeneity and latent space heterogeneity 
among domains, which limits the application of existing methods and even leads to 
negative transfer.

\section{Model Description\label{Model Description}}
\subsection{Architecture Overview}
\begin{figure*}[h]
    \centering
    \includegraphics[width=\linewidth]{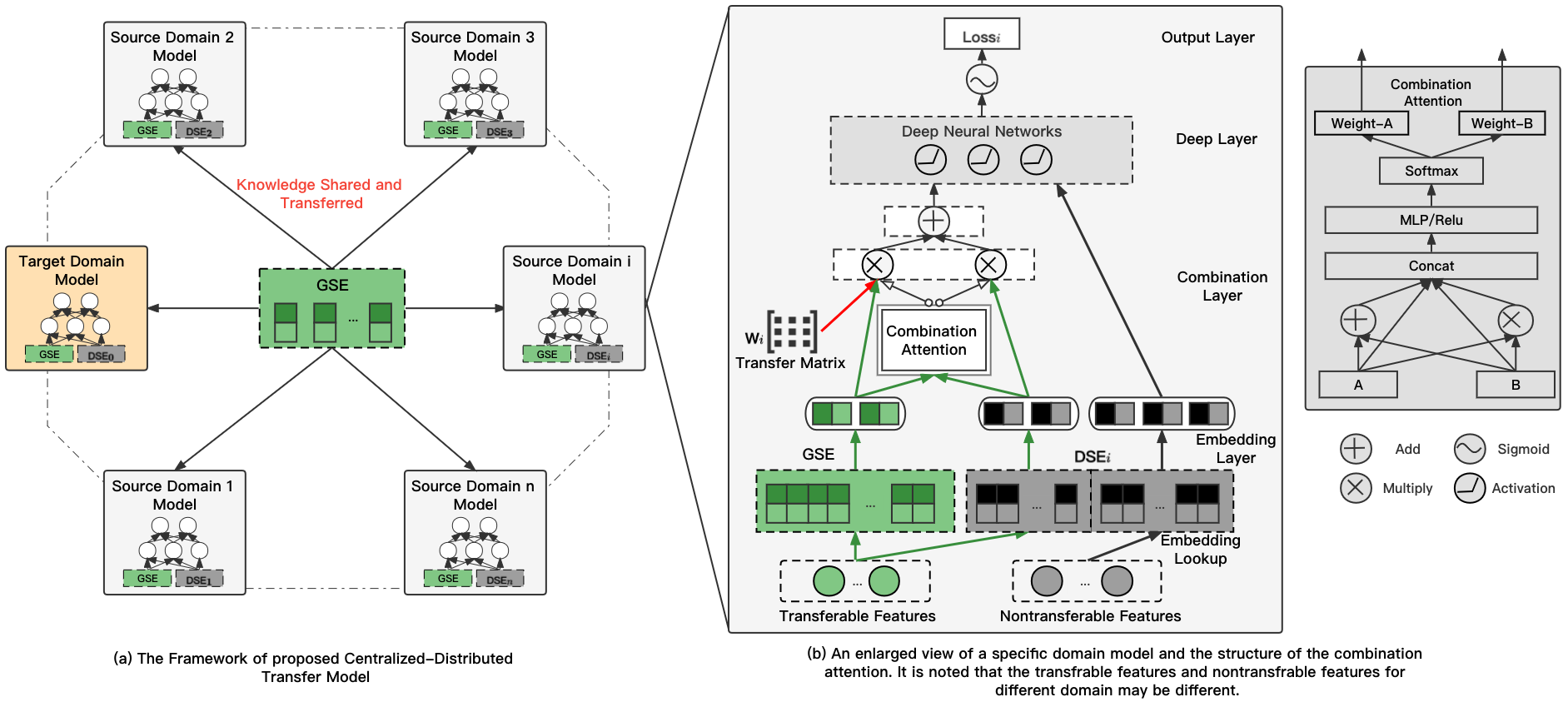}
    \caption{The Framework  of proposed Centralized-Distributed Transfer Model
    (GSE ie. Global shared embedding, DSE ie. Domain specific embedding. The Deep neural networks for each domain can be any single-domain model, such as DeepFM,DIN,etc.)
    }
    \label{fig2}
\end{figure*}

In this section, we present a centralized-distributed transfer model for CDR based on multi 
source heterogeneous transfer learning, called CDTM. As shown in Fig.~\ref{fig2}(a), there 
are one target domain model and several source domain models distributed around the 
framework. Each model has its own DSE with the domain number. In the center position, 
the GSE is shared by all the models. The Fig.~\ref{fig2}(b) illustrate 
an enlarged view of a specific domain model, which is 
divided into four main components, i.e., Embedding Layer, Combination Layer, 
Deep Layer, and Output Layer.

\subsection{Embedding Layer\label{Embedding Layer}}
\subsubsection{Dual Embedding Structure}
\ 
\newline 
In order to effectively transfer multi-source information to 
the target domain, we propose the dual embedding structure: DSE 
trained separately by each domain, and GSE trained by all domains jointly.

The $q+1$ domains are denoted as ${s_i}\left(i\in\left[0,q\right]\right)$, where $s_0$ is the target domain 
and the others are source domains. For domain $s_i$, its input feature vector 
is $\mathbf{X}^i\in\mathbb{R}^{f_i\times1}=[\mathbf{X}_c^i,\mathbf{X}_d^i]$, where $f_i$ 
is the number of feature fields. $\mathbf{X}_c^i\in\mathbb{R}^{m_i\times1}$ represents 
the transferable feature vector and $m_i$ is the transferable feature field number. 
$\mathbf{X}_d^i\in\mathbb{R}^{(f_i-m_i)\times1}$ is the nontransferable feature vector, 
where the feature fields are unique to $s_i$.

The transferable features have only one type of 
embedding (i.e., DSE), while the nontransferable features have two (DSE and GSE).
For nontransferable feature vector $\mathbf{X}_d^i$, its corresponding embedding 
$\mathbf{E}_d^i\in\mathbb{R}^{(f_i-m_i))\times k}$ can be obtained by looking 
up the DSE table $\mathbf{W}^i\in\mathbb{R}^{n_i\times k}$, where $k$ is the 
dimension of embedding, and $n_i$ is the number of features. For the transferable 
feature vector $\mathbf{X}_c^i$, both DSE table $\mathbf{W}^i$ and GSE table 
$\mathbf{W}^g\in\mathbb{R}^{p\times k}$(p is the number of shared features) 
will be looked up to obtain $\mathbf{E}_c^i\in\mathbb{R}^{m_i\times k}$ and 
$\mathbf{G}_c^i\in\mathbb{R}^{m_i\times k}$ respectively. Therefore, the feature 
embeddings of domain $s_i$ consist of three parts: $\mathbf{E}_d^i,\mathbf{E}_c^i$ 
and $\mathbf{G}_c^i$.

With the dual embedding component, domain $s_i$ can use DSE to represent its unique characteristics, 
and at the same time, it can also obtain global feature representation through GSE trained jointly by all domains.

\subsection{Combination Layer}
% As mentioned in Section 3.2, for a given domain $s_i\left(i\in\left[0,q\right]\right)$, there are two 
% embeddings for its transfer features: domain specific embedding $\mathbf{E}_c$ and global 
% shared embedding $\mathbf{G}_c$. For each feature embedding $\mathbf{e}_c^j\in\mathbb{R}^{1\times k}$ 
% in $\mathbf{E}_c$, the corresponding global shared embedding is denoted 
% as $\mathbf{g}_c^j\in\mathbb{R}^{1\times k}$. The combine layer combines these two 
% embeddings together by the following formula:
As mentioned in Section~\ref{Embedding Layer}, for a given domain $s_i$, there are two 
embeddings for its transferable features: domain specific embedding $\mathbf{E}_c$ and global 
shared embedding $\mathbf{G}_c$. The combination layer combines them together as follows:
% \begin{eqnarray}
%   \mathbf{e}^j=\mathbf{e}_c^j\otimes \mathbf{A}^j+\mathbf{T}^j\otimes\mathbf{g}_c^j\otimes(1-\mathbf{A}^j)
% \end{eqnarray}
\begin{align}
\mathbf{E}=\mathbf{E}_c\otimes \mathbf{A}+\mathbf{T}\otimes\mathbf{G}_c\otimes(1-\mathbf{A})
\end{align}
% where $\mathbf{T}^j$ is the transfer matrix used to map the GSE to the latent space of DSE. 
% $\mathbf{A}^j$ is the combine attention matrix, which is used to effectively combine GSE and DSE. 
% Next, we will introduce transfer matrix and combine attention in detail.
where $\mathbf{T}$ is the transfer matrix to map the GSE to the latent space of DSE. 
$\mathbf{A}$ is the combination attention matrix, which is used to effectively combine GSE and DSE. 
Next, we will introduce transfer matrix and combination attention in detail.

\subsubsection{Transfer Matrix}
\ 
\newline 
% To deal with the latent space heterogeneity of DSE and GSE, the transfer matrix is proposed 
% to map them into a shared latent space. The optimization goal is to search for a transfer 
% matrix that minimizes the Euclidean distance between the domain specific embedding $\mathbf{e}_c^j$ 
% and global shared embedding  $\mathbf{g}_c^j$.
To deal with the latent space heterogeneity of DSE and GSE, the transfer matrix is proposed 
to map them into a shared latent space. The optimization goal is to search for a transfer 
matrix that minimizes the Euclidean distance between the domain specific embedding $\mathbf{E}_c$ 
and global shared embedding  $\mathbf{G}_c$.
% \begin{eqnarray}
% {L}^j=argmin_{\mathbf{T}^j}\sum{\parallel\mathbf{e}_c^j-\mathbf{T}^j\otimes \mathbf{g}_c^j\parallel}^2
% \end{eqnarray}
% \begin{align}
% {L}^j=argmin_{\mathbf{T}^j}\sum{\parallel\mathbf{e}_c^j-\mathbf{T}^j\otimes \mathbf{g}_c^j\parallel}^2
% \end{align}
\begin{align}
    {L}=\underset{\mathbf{T}}{argmin}\sum{\parallel\mathbf{E}_c-\mathbf{T}\otimes \mathbf{G}_c\parallel}^2
\end{align}
This transformation maintains two facts: (i) DSE and GSE should be different 
because they keep different information; (ii) the mapped 
GSE should be in the shared latent space with DSE.

\subsubsection{Combination Attention}
\ 
\newline 
Note that the same feature is of different importance in different domains. To model this importance, 
the combination attention is applied to weight DSE and GSE.

% For a transfer feature, its DSE and corresponding GSE are $\mathbf{e}_c^j$ and $\mathbf{g}_c^j$ respectively. 
% Formally, we define the Combine Attention as follows:
For a transferable feature, its DSE and corresponding GSE are $\mathbf{E}_c$ and $\mathbf{G}_c$ respectively. 
Formally, we define the combination attention as follows:
% \begin{eqnarray}
% \mathbf{h}_0=activation\left(\mathbf{V}_0\left[\mathbf{e}_c^j,\mathbf{e}_c^j\otimes\mathbf{g}_c^j,\mathbf{e}_c^j\oplus\mathbf{g}_c^j,\mathbf{g}_c^j\right]+\mathbf{b}_0\right)
% \end{eqnarray}
% \begin{align}
% \mathbf{h}_0=Relu\left(\mathbf{V}_0\left[\mathbf{e}_c^j,\mathbf{e}_c^j\otimes\mathbf{g}_c^j,\mathbf{e}_c^j\oplus\mathbf{g}_c^j,\mathbf{g}_c^j\right]+\mathbf{b}_0\right)
% \end{align}
\begin{align}
\mathbf{h}_0=Relu\left(\mathbf{V}_0\left[\mathbf{E}_c,\mathbf{E}_c\otimes\mathbf{G}_c,\mathbf{E}_c\oplus\mathbf{G}_c,\mathbf{G}_c\right]
+\mathbf{b}_0\right)
\end{align}
% \begin{eqnarray}
% \mathbf{h}_1=activation\left(\mathbf{V}_1\mathbf{h}_0+\mathbf{b}_1\right)
% \end{eqnarray}
% \begin{align}
% \mathbf{h}_1=Relu\left(\mathbf{V}_1\mathbf{h}_0+\mathbf{b}_1\right)
% \end{align}
% \begin{eqnarray}
%     \mathbf{A}^j=\sigma(\mathbf{V}_2\mathbf{h}_1+\mathbf{b}_2)
% \end{eqnarray}
\begin{align}
\mathbf{A}=\sigma(\mathbf{V}_1\mathbf{h}_0+\mathbf{b}_1)
\end{align}
% where $\otimes$ and $\oplus$ represent the element wise multiplication and element wise addition, 
% respectively. $\sigma(\bullet)$ and $Relu(\bullet)$ are the sigmoid function and the Relu\cite{glorot2011deep}
% activation function, respectively. $\mathbf{V}_0,\mathbf{V}_1,\mathbf{V}_2$ are the hidden layer 
% weights. And $\mathbf{b}_0,\mathbf{b}_1,\mathbf{b}_2$ are the biases. Note that $\mathbf{A}^j$ is a vector used to weight 
% DSE $\mathbf{e}_c^j$, while the GSE $\mathbf{g}_c^j$ mapped with the transfer matrix $\mathbf{T}^j$ is 
% weighted by $1-\mathbf{A}^j$.
where $\otimes$ and $\oplus$ represent the element wise multiplication and element wise addition, 
respectively. $\sigma(\bullet)$ and $Relu(\bullet)$ are the sigmoid function and the relu
activation function, respectively. $\mathbf{V}_0,\mathbf{V}_1$ are the hidden layer 
weights. And $\mathbf{b}_0,\mathbf{b}_1$ are the biases. Note that $\mathbf{A}$ is a vector used to weight 
$\mathbf{E}_c$, while $\mathbf{G}_c$ mapped with the transfer matrix $\mathbf{T}$ is 
weighted by $1-\mathbf{A}$.

It is noted that the combined embedding of all the transferable features $\mathbf{E}$ 
is one of the inputs of the deep layer, and the other one 
is the embeddings of nontransferable features $\mathbf{E}_d$. In this way, useful information 
from $\mathbf{W}^g$ jointly trained by all domains is transferred into the target domain to 
improve its model. Recall that the transferable feature fields of all domains are not the same, 
because there is feature dimension heterogeneity between the target domain and different source domains. 
Only the common feature fields of the source domain and the target domain will be transferred.

\subsection{Deep Layer and Output Layer}
In deep layer, the output of combination layer, denoted by $\mathbf{x}_0=[\mathbf{E},\mathbf{E}_d]$, 
is applied to two fully connected layers with relu activation function to obtain 
nonlinear relationships between features. Then, the output of the last deep layer, denoted 
by $\mathbf{o}$ is fed into an output layer with sigmoid activation to get the model prediction.
\begin{align}
p=\sigma(\mathbf{Q}^T\mathbf{o}+z)
\end{align}
where $\mathbf{Q}^T$ are the hidden layer weights, 
$\mathbf{z}$ are the biases, $p$ is the model prediction.

\subsection{Model Training and Auxiliary Loss}
The model proposed in this paper is jointly trained by multiple domains. Each model trains its own DSE 
independently, and at the same time, all the models train GSE jointly. The prediction loss 
function for a specific domain is defined as:
% \begin{eqnarray}
% Loss_i=-\frac{1}{|Y_i|}\sum_{y_i\in Y_i}{[y_ilog(p_i)+(1-y_i)log(1-p_i)]}
% \end{eqnarray}
\begin{align}
Loss_i=-\frac{1}{|Y_i|}\sum_{y_i\in Y_i}{[y_ilog(p_i)+(1-y_i)log(1-p_i)]}
\end{align}

In addition, an auxiliary loss is constructed based on the previously described 
optimization objective of transfer matrix:
% \begin{eqnarray}
% L_i=\lambda\sum{\parallel\mathbf{e}_c^j-\mathbf{W}^j\mathbf{g}_c^j\parallel}^2
% \end{eqnarray}
% \begin{align}
% L_i=\lambda\sum{\parallel\mathbf{e}_c^j-\mathbf{T}^j\otimes \mathbf{g}_c^j\parallel}^2
% \end{align}
\begin{align}
L^*_i=\lambda\sum{\parallel\mathbf{E}_c-\mathbf{T}\otimes \mathbf{G}_c\parallel}^2
\end{align}
where $\lambda$ is the regularization parameter.
In conclusion, the total loss of the model can be obtained:
% \begin{eqnarray}
% Loss=\sum_{i=0}^{q}{\alpha_i(Loss_i+\lambda L_i)}
% \end{eqnarray}
\begin{align}
Loss=\sum_{i=0}^{q}{\alpha_i(Loss_i+\lambda L^*_i)}
\end{align}
where $\alpha_i$ is the balance coefficient of the loss function.

\subsection{Symmetrical Structure and Extensibility}
As described before, it can be easily concluded that the proposed CDTM model structure is symmetrical. 
In other words, any domain can be regarded as the target domain, and the others as the source domains. 
For a specific domain, it receives information transferred from other domains to improve its model, 
at the same time, this domain also contributes to the optimization of other domains as a source.

% Therefore, the CDTM model in this paper can be regarded as a paradigm for 
% solving multi-source cross-domain recommendation. And users can customize the specific model 
% architecture according to their scenarios: it can be single-source or multi-source; 
% it can be single-target cross-domain recommendation, 
% dual-target cross-domain recommendation or multi-target. For each specific domain, 
% users can also customize the deep network structure such as DCN, DIN, etc.

\section{Experiments\label{Experiments}}
To verify the effectiveness of the model proposed, we conduct extensive experiments on real industrial data. 
% The design and implementation of the experiments focus on answering following questions:
% \ 
% \newline 
% Q1: Can the proposed model handle the cross-domain recommendation problem in multi-source heterogeneous scenarios?
% \ 
% \newline 
% Q2: Does the proposed model perform better than other methods in different cross-domain recommendation scenarios?
% \ 
% \newline 
% Q3: Can the proposed method achieve better model performance through multi-source transfer than single-source transfer?
% \ 
% \newline 
% Q4: How do the proposed dual embedding and combine attention components contribute to the performance of the proposed model?
% \ 
% \newline 
% Q5: Whether the proposed model can be extended to multi-domain and multi-target scenarios, where each domain is both
% the source and target for other domain, to simultaneously improve the performance of multiple domain models.

\subsection{Experimental Setup}
\subsubsection{Dataset}
\ 
\newline 
Due to the lack of suitable open public datasets for multi-source CDR, we choose the 
real-world commercial data sampled from the NetEase Cloud Music advertising system as the experiment 
dataset. The description of the dataset is shown in Table~\ref{Table1}. As shown in Table~\ref{Table1}, 
compared with the F1-F4 domains, H and J domains have more sufficient data. 
Besides, H domain is a contract ads flight while others are bid ads flight. In our experiments, 
H and J domains are regarded as the source domains, while the F1-F4 domains are regarded as the target domains. 
And it can be found that the transferable features of these domains are different. 
There are obvious latent space heterogeneity and feature dimensional heterogeneity 
between the source and target domains.

\begin{table}[htbp]
    \centering
    \begin{threeparttable}
    \caption{Dataset description}
    \label{Table1}
    \renewcommand\arraystretch{1.2} 
    \arrayrulecolor{black}
    \begin{tabular}{lllllll} 
    \hline
    Domain & H & J & F1 & F2 & F3 & F4 \\ 
    \hline
    Data size(G) & 450 & 400 & 22 & 39 & 62 & 53 \\
    Feature field & 527 & 613 & 603 & 603 & 603 & 603 \\
    TFF number\tnote{1} & 386 & 555 & 555 & 555 & 555 & 555 \\
    CTR(\%) & 1.29 & 0.61 & 0.15 & 0.21 & 0.41 & 0.17 \\
    ads type & contract & bid & bid & bid & bid & bid \\
    \hline
    \end{tabular}
    \arrayrulecolor{black}
    \begin{tablenotes}
        \item[1] FFT is the abbreviation of transferable feature field
    \end{tablenotes}
\end{threeparttable}
\end{table}

\subsubsection{experimental Tasks}
\ 
\newline
The following 4 tasks are designed:

Task1: This task takes H/J domain as the source domain. For each target domain in F1/F2/F3/F4, 
we will train a model separately to verify the single-source CDR
performance of the proposed model in different target scenarios.

% Task2: This task takes J domain as the source domain and F1/F2/F3/F4 as the target domain to 
% verify the single-source cross-domain recommendation performance of the proposed model. 
% Similarly, models will be trained separately for each target domain.

Task2: This task takes both H and J as source domains, while F1/F2/F3/F4 as target domains 
to verify whether the proposed model can achieve better target model performance through 
multi-source transfer than single-source transfer.

Task3: In this task, we conduct an ablation study to examine how the dual embedding 
and combine attention contribute to the performance of the proposed model.

Task4: In this task, the proposed model CDTM is jointly trained by the 4 domains(F1/F2/F3/F4). 
And each domain can be regarded as the source domain or target domain. This task is designed 
to verify whether the proposed model can be extended to MDR scenarios.

\subsubsection{Compared Models}
\ 
\newline 
We compare our CDTM model with these recently proposed models: 
CoNeT\cite{hu2018conet},SCoNet\cite{hu2018conet},
DDTCDR\cite{li2020ddtcdr},DTCDR\cite{zhu2019dtcdr} and GA-DTCDR\cite{zhu2021unified}, to demonstrate 
the superiority of the proposed model. The single domain model is a DCN\cite{wang2017deep} model 
trained by each domain using its own domain data.

For all models except for Base, the additional "-(Domain)" suffix indicates the source domain used during 
training, and the no-suffix indicates the model using both H and J domains as the sources. 
For example, CoNet-H represents a CoNet model trained using H domain as the source domain, 
and CDTM represents a model trained using both H and J domains as source domains.

To obtain a fair comparison, the deep layers of all the models are 2-layer fully-connected 
networks. The hidden layer size are 200,128. The activation for the hidden layers is relu. 
The regularization parameter is set to 0.0001. All the models 
are optimized using adam algorithm with a learning rate 0.001.

\subsubsection{Evaluation Metrics}
\ 
\newline 
\textbf{AUC:} AUC\cite{ouyang2020minet} is widely used in the 
evaluation of CTR models. The larger the AUC is, the better the model performs. Even a small improvement 
in AUC can lead to a significant improvement in online performance.
\ 
\newline 
\textbf{Imp:} The relative improvement of the specific model AUC over the Base model. It is defined as:
% \begin{eqnarray}
% Imp=\frac{AUC-AUC_t}{AUC_t}\times100%
% \end{eqnarray}
\begin{align}
Imp=\frac{AUC-AUC_t}{AUC_t}\times100%
\end{align}
where $AUC_t$ represents the AUC of Base model and $AUC$ is the specific model AUC. 
The larger the Imp, the greater the improvement of the AUC of the model relative to the base model.

\subsection{Experimental Results}
\subsubsection{Result 1: Performance Comparison (for Task1)}
\ 
\newline
\indent As shown in Table~\ref{Table2} (take H domain as the source domain), 
the proposed CDTM model achieves the best results on different target domain. 
For the F1/F2/F3/F4 target domain, our CDTM model improves Base by 3.62\%, 0.58\%, 1.43\% and 0.64\% in terms 
of AUC, respectively while the improvements of the best baseline for AUC are 1.54\%, 0.12\%, 1.12\% and 0.54\%, 
respectively. The results that take J domain as the source domain shown in Table~\ref{Table3} are similar.
This indicates the proposed CDTM can obtain better transfer results than the other models. In addition, 
note that for the four target domains, except for our CDTM model, other models have different degrees 
of negative transfer phenomenon, which may be caused by the heterogeneous difference between sources and 
targets. This also demonstrates the effectiveness and robustness of our 
model for heterogeneous CDR.

\begin{table*}[htbp]
    \caption{Comparison of the results of different methods for Task 1}
    \label{Table2}
    \renewcommand\arraystretch{1.2} 
    \centering
    \arrayrulecolor{black}
    \begin{tabular}{cccccccccccccc}
    \hline
    \multirow{2}{*}{Flight} & Base & \multicolumn{2}{c}{CoNet-H} & \multicolumn{2}{c}{SCoNet-H} & \multicolumn{2}{c}{DDTCDR-H} & \multicolumn{2}{c}{DTCDR-H} & \multicolumn{2}{c}{GA-DTCDR-H} & \multicolumn{2}{c}{CDTM-H} \\ 
    \cline{2-14}
     & AUC & AUC & Imp & AUC & Imp & AUC & Imp & AUC & Imp & AUC & Imp & AUC & Imp \\ 
    \hline
    F1 & 0.5778 & 0.5836 & 1.00\% & 0.5814 & 0.62\% & 0.5805 & 0.47\% & 0.5842 & 1.11\% & 0.5867* & 1.54\%* & \textbf{0.5987} & \textbf{3.62\%} \\
    F2 & 0.6049 & 0.6051 & 0.03\% & 0.6054 & 0.08\% & 0.6056* & 0.12\%* & 0.6047 & -0.03\% & 0.6042 & -0.12\% & \textbf{0.6084} & \textbf{0.58\%} \\
    F3 & 0.5893 & 0.5932 & 0.66\% & 0.5899 & 0.10\% & 0.5892 & -0.02\% & 0.5947 & 0.92\% & 0.5959* & 1.12\%* & \textbf{0.5977} & \textbf{1.43\%} \\
    F4 & 0.5965 & 0.5839 & -2.11\% & 0.5922 & -0.72\% & 0.5954 & -0.18\% & 0.5999* & 0.57\%* & 0.5997 & 0.54\% & \textbf{0.6021} & \textbf{0.94\%} \\
    \hline
    \end{tabular}
    \arrayrulecolor{black}
\end{table*}

\begin{table*}[htbp]
    \caption{Comparison of the results of different methods for Task 2}
    \label{Table3}
    \renewcommand\arraystretch{1.2} 
    \centering
    \arrayrulecolor{black}
    \begin{tabular}{cccccccccccccc} 
    \hline
    \multirow{2}{*}{Flight} & Base & \multicolumn{2}{c}{CoNet-J} & \multicolumn{2}{c}{SCoNet-J} & \multicolumn{2}{c}{DDTCDR-J} & \multicolumn{2}{c}{DTCDR-J} & \multicolumn{2}{c}{GA-DTCDR-J} & \multicolumn{2}{c}{CDTM-J} \\ 
    \cline{2-14}
     & AUC & AUC & Imp & AUC & Imp & AUC & Imp & AUC & Imp & AUC & Imp & AUC & Imp \\ 
    \hline
    F1 & 0.5778 & 0.5803 & 0.43\% & 0.5925* & 2.54\%* & 0.5796 & 0.31\% & 0.5853 & 1.30\% & 0.5864 & 1.49\% & \textbf{0.5965} & \textbf{3.24\%} \\
    F2 & 0.6049 & 0.6051* & 0.03\%* & 0.6046 & -0.05\% & 0.6039 & -0.17\% & 0.6039 & -0.17\% & 0.6047 & -0.03\% & \textbf{0.6091} & \textbf{0.69\%} \\
    F3 & 0.5893 & 0.5856 & -0.63\% & 0.5825 & -1.15\% & 0.5884 & -0.15\% & 0.5947 & 0.92\% & 0.5948* & 0.93\%* & \textbf{0.5959} & \textbf{1.12\%} \\
    F4 & 0.5965 & 0.5826 & -2.33\% & 0.5947 & -0.30\% & 0.5884 & -1.36\% & 0.5994 & 0.49\% & 0.6019* & 0.91\%* & \textbf{0.6024} & \textbf{0.99\%} \\
    \hline
    \end{tabular}
    \arrayrulecolor{black}
\end{table*}

\subsubsection{Result 2: Multi-Source vs Single-Source (for Task2)}
\ 
\newline
\indent To demonstrate the proposed CDTM can achieve better results through multi-source transfer 
than single-source transfer, we compare CDTM with CDTM-H and the CDTM-J, 
then summarize all the results in Table~\ref{Table4}.

\begin{table}[htbp]
    \caption{Comparison of the results of for Task 3}
    \label{Table4}
    \renewcommand\arraystretch{1.2} 
    \centering
    \arrayrulecolor{black}
    \begin{tabular}{ccccccc} 
    \hline
    \multirow{2}{*}{Flight} & \multicolumn{2}{c}{CDTM-H} & \multicolumn{2}{c}{CDTM-J} & \multicolumn{2}{c}{CDTM} \\ 
    \cline{2-7}
     & AUC & Imp & AUC & Imp & AUC & Imp \\ 
    \hline
    F1 & 0.5987 & 3.62\% & 0.5965 & 3.24\% & \textbf{0.6042} & \textbf{4.57\%} \\
    F2 & 0.6084 & 0.58\% & 0.6091 & 0.69\% & \textbf{0.6121} & \textbf{1.19\%} \\
    F3 & 0.5977 & 1.43\% & 0.5959 & 1.12\% & \textbf{0.5989} & \textbf{1.63\%} \\
    F4 & 0.6021 & 0.94\% & 0.6024 & 0.99\% & \textbf{0.6044} & \textbf{1.32\%} \\
    \hline
    \end{tabular}
    \arrayrulecolor{black}
\end{table}

As illustrated in Table~\ref{Table4}, CDTM performs better than CDTM-H and CDTM-J on all target domains, 
indicating that the proposed model can effectively utilize multi-source information and achieve better 
results for multi-source transfer than single-source transfer. Recall that the source domain and the 
target domain are heterogeneous, so the experimental results further verify the effectiveness 
and robustness of our CDTM model for multi-source heterogeneous cross-domain transfer.

% \subsection{Ablation Study}
% As mentioned above, the proposed CDTM model achieves significant improvements over other baseline models, 
% mainly due to two novel model structure designs: The first is dual embedding structure, 
% which utilizes DSE and GSE to maintain the feature representation of specific domain 
% and global feature distribution, respectively. The second is the
% combination attention, which calculates the importance of DSE and GSE adaptively.
\subsubsection{Result 3: Ablation Study Result (for Task3)}
\begin{figure}[h]
    \centering
    \includegraphics[width=\linewidth]{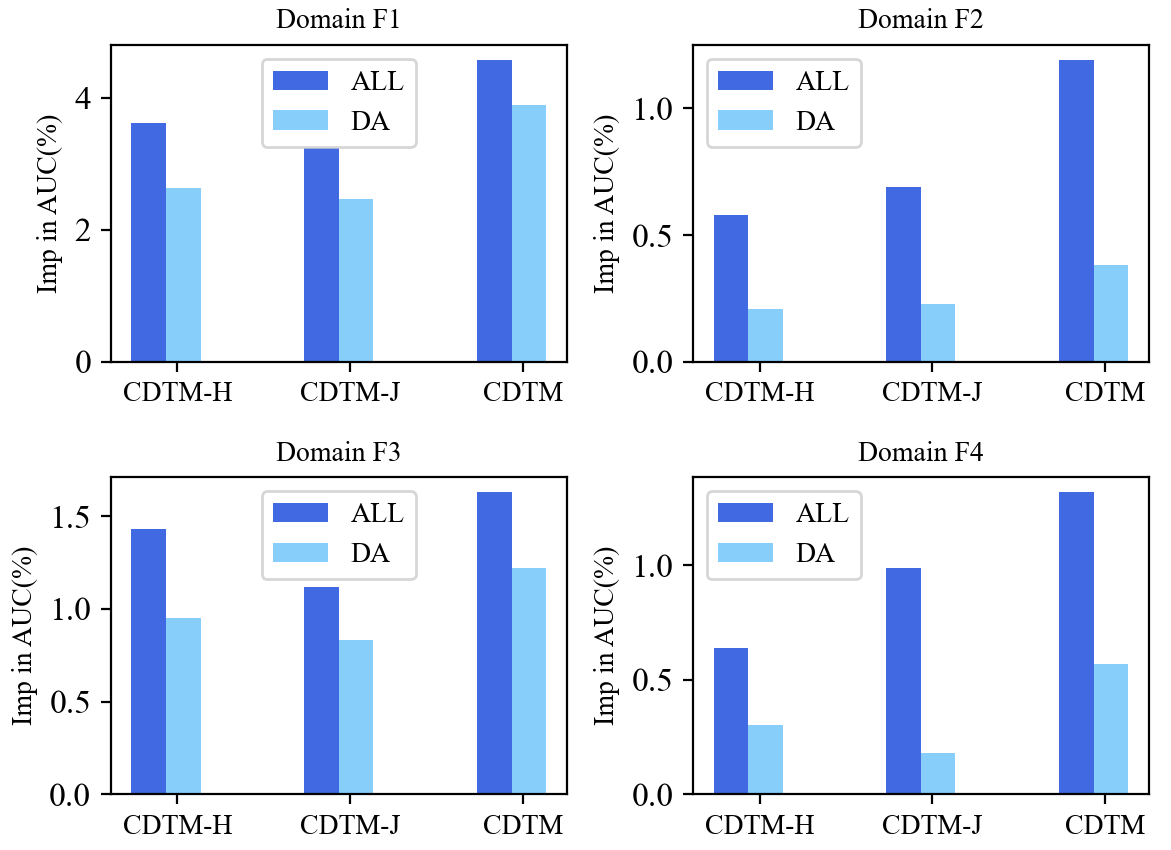}
    \caption{Comparison of ablation experiments (DA represents the model that 
    removes combination attention, and ALL is the original model)}
    \label{fig3}
\end{figure}

\ 
\newline
\indent We conduct ablation experiments to investigate the contributions of dual embedding 
structure and combine attention. The CDTM model that drops combination attention is denoted as CDTM-DA. 
The ablation study results are illustrated in Fig~\ref{fig3}. 

% \begin{table*}[htbp]
%     \caption{Comparison of results of ablation experiment in Task 4}
%     \label{Table5}
%     \renewcommand\arraystretch{1.2} 
%     \centering
%     \arrayrulecolor{black}
%     \begin{tabular}{cccccccccccccc} 
%     \hline
%     \multirow{2}{*}{Flight} & Base & \multicolumn{2}{c}{CDTM-DA-H} & \multicolumn{2}{c}{CDTM-H} & \multicolumn{2}{c}{CDTM-DA-J} & \multicolumn{2}{c}{CDTM-J} & \multicolumn{2}{c}{CDTM-DA} & \multicolumn{2}{c}{CDTM} \\ 
%     \cline{2-14}
%      & AUC & AUC & Imp & AUC & Imp & AUC & Imp & AUC & Imp & AUC & Imp & AUC & Imp \\ 
%     \hline
%     F1 & 0.5778 & 0.5930 & 2.63\% & 0.5987 & 3.62\% & 0.5920 & 2.46\% & 0.5965 & 3.24\% & 0.6003 & 3.89\% & 0.6042 & 4.57\% \\
%     F2 & 0.6049 & 0.6062 & 0.21\% & 0.6084 & 0.58\% & 0.6063 & 0.23\% & 0.6091 & 0.69\% & 0.6072 & 0.38\% & 0.6121 & 1.19\% \\
%     F3 & 0.5893 & 0.5949 & 0.95\% & 0.5977 & 1.43\% & 0.5942 & 0.83\% & 0.5959 & 1.12\% & 0.5965 & 1.22\% & 0.5989 & 1.63\% \\
%     F4 & 0.5965 & 0.5983 & 0.30\% & 0.6021 & 0.94\% & 0.5976 & 0.18\% & 0.6024 & 0.99\% & 0.5999 & 0.57\% & 0.6044 & 1.32\% \\
%     \hline
%     \end{tabular}
%     \arrayrulecolor{black}
% \end{table*}

As shown in Fig~\ref{fig3}, the CDTM-DA model obtains better results than Base model in all target domain experiments, 
indicating that the design of dual embedding structure does improve the effect of cross-domain transfer. Besides, 
it can be seen from Fig~\ref{fig3} that in all domains, CDTM performs significantly than CDTM-DA. This shows that the design 
of combine attention does further improve the model performance.

% \subsection{Extensibility Study}
\subsubsection{Result 4: Extensibility Study Result (for Task4)}
% \subsubsection{Result 5: Extensibility Study Result (for Task5)}
\ 
\newline
\indent In task 4, the CDTM model is jointly trained by F1/F2/F3/F4 and we denote this model as CDTM$_4$. 
The comparison results between CDTM$_4$ and Base model are shown in Table~\ref{Table6}.

As shown in Table~\ref{Table6}, the CDTM$_4$ model achieves AUC improvements by 3.28\%, 0.41\%, 
0.85\% and 0.63\% for the F1-F4 domains, respectively. Therefore, our CDTM 
can improve the performance of multiple target domains simultaneously, which is mainly 
due to the symmetrical centralized-distributed structure design and advanced scalability 
of our model. This demonstrate that the propose CDTM can 
be extended to multi-source and multi-target CDR.

\begin{table}[htbp]
    \caption{Extensibility study result for task 4}
    \label{Table6}
    \renewcommand\arraystretch{1.2} 
    \centering
    \arrayrulecolor{black}
    \begin{tabular}{cccc} 
    \hline
    \multirow{2}{*}{Flight} & Base & \multicolumn{2}{c}{CDTM$_4$} \\ 
    \cline{2-4}
     & AUC & AUC & Imp \\ 
    \hline
    F1 & 0.5778 & 0.6005 & 3.93\% \\
    F2 & 0.6049 & 0.6089 & 0.66\% \\
    F3 & 0.5893 & 0.5956 & 1.07\% \\
    F4 & 0.5965 & 0.6009 & 0.74\% \\
    \hline
    \end{tabular}
    \arrayrulecolor{black}
\end{table}

\section{Online A/B Test\label{Online A/B Test}}
We deploy our CDTM model to the NetEase Cloud Music advertising A/B Test system. 
During a two-week online A/B Test, we evaluated the CTR and effective cost per mille (eCPM) 
of the CDTM model and the baseline model (DCN). 
Online A/B Test results show that our CDTM model achieves a 5.1\% improvement 
in CTR and a 6.6\% improvement in eCPM relative to the baseline model, which 
demonstrates the effectiveness of the model in CDR. Currently, 
the CDTM model has been deployed in our online advertising system.
% NetEase Cloud Music

\section{Conclusions\label{Conclusions}}
In this paper, we proposed a centralized-distributed transfer model for CDR based on multi-source 
heterogeneous transfer learning, which can be customized and extended 
to more domain scenarios. The dual embedding structure, which includes DSE trained 
by each domain and trained jointly by all the domains, is constructed to generate more representative 
feature representation. The transfer matrix is utilized to map the GSE to the feature 
space of the target domain, and an auxiliary loss is constructed to help the 
optimization of the transfer matrix. Then, the combination attention is utilized to 
adaptively combine GSE and DSE of the transfer features. Extensive offline and online experiment 
results on industrial datasets demonstrate the effectiveness, robustness, and 
extensibility of our model. In cross-domain recommendation, the user interest 
transfer is also very important. In future work, 
we will investigate the user behavior sequence transfer.

\bibliographystyle{IEEEtran}
\bibliography{IEEEabrv,cite.bib}

\begin{thebibliography}{10}
\providecommand{\url}[1]{#1}
\csname url@samestyle\endcsname
\providecommand{\newblock}{\relax}
\providecommand{\bibinfo}[2]{#2}
\providecommand{\BIBentrySTDinterwordspacing}{\spaceskip=0pt\relax}
\providecommand{\BIBentryALTinterwordstretchfactor}{4}
\providecommand{\BIBentryALTinterwordspacing}{\spaceskip=\fontdimen2\font plus
\BIBentryALTinterwordstretchfactor\fontdimen3\font minus
  \fontdimen4\font\relax}
\providecommand{\BIBforeignlanguage}[2]{{%
\expandafter\ifx\csname l@#1\endcsname\relax
\typeout{** WARNING: IEEEtran.bst: No hyphenation pattern has been}%
\typeout{** loaded for the language `#1'. Using the pattern for}%
\typeout{** the default language instead.}%
\else
\language=\csname l@#1\endcsname
\fi
#2}}
\providecommand{\BIBdecl}{\relax}
\BIBdecl

\bibitem{guo2017deepfm}
H.~Guo, R.~Tang, Y.~Ye, Z.~Li, and X.~He, ``Deepfm: a factorization-machine
  based neural network for ctr prediction,'' \emph{arXiv preprint
  arXiv:1703.04247}, 2017.

\bibitem{zhou2018deep}
G.~Zhou, X.~Zhu, C.~Song, Y.~Fan, H.~Zhu, X.~Ma, Y.~Yan, J.~Jin, H.~Li, and
  K.~Gai, ``Deep interest network for click-through rate prediction,'' in
  \emph{Proceedings of the 24th ACM SIGKDD International Conference on
  Knowledge Discovery \& Data Mining}, 2018, pp. 1059--1068.

\bibitem{zang2021survey}
T.~Zang, Y.~Zhu, H.~Liu, R.~Zhang, and J.~Yu, ``A survey on cross-domain
  recommendation: Taxonomies, methods, and future directions,'' \emph{arXiv
  preprint arXiv:2108.03357}, 2021.

\bibitem{richardson2007predicting}
M.~Richardson, E.~Dominowska, and R.~Ragno, ``Predicting clicks: estimating the
  click-through rate for new ads,'' in \emph{Proceedings of the 16th
  international conference on World Wide Web}, 2007, pp. 521--530.

\bibitem{rendle2010factorization}
S.~Rendle, ``Factorization machines,'' in \emph{2010 IEEE International
  conference on data mining}.\hskip 1em plus 0.5em minus 0.4em\relax IEEE,
  2010, pp. 995--1000.

\bibitem{juan2016field}
Y.~Juan, Y.~Zhuang, W.-S. Chin, and C.-J. Lin, ``Field-aware factorization
  machines for ctr prediction,'' in \emph{Proceedings of the 10th ACM
  conference on recommender systems}, 2016, pp. 43--50.

\bibitem{qu2016product}
Y.~Qu, H.~Cai, K.~Ren, W.~Zhang, Y.~Yu, Y.~Wen, and J.~Wang, ``Product-based
  neural networks for user response prediction,'' in \emph{2016 IEEE 16th
  International Conference on Data Mining (ICDM)}.\hskip 1em plus 0.5em minus
  0.4em\relax IEEE, 2016, pp. 1149--1154.

\bibitem{cheng2016wide}
H.-T. Cheng, L.~Koc, J.~Harmsen, T.~Shaked, T.~Chandra, H.~Aradhye,
  G.~Anderson, G.~Corrado, W.~Chai, M.~Ispir \emph{et~al.}, ``Wide \& deep
  learning for recommender systems,'' in \emph{Proceedings of the 1st workshop
  on deep learning for recommender systems}, 2016, pp. 7--10.

\bibitem{wang2017deep}
R.~Wang, B.~Fu, G.~Fu, and M.~Wang, ``Deep \& cross network for ad click
  predictions,'' in \emph{Proceedings of the ADKDD'17}, 2017, pp. 1--7.

\bibitem{cheng2020adaptive}
W.~Cheng, Y.~Shen, and L.~Huang, ``Adaptive factorization network: Learning
  adaptive-order feature interactions,'' in \emph{Proceedings of the AAAI
  Conference on Artificial Intelligence}, vol.~34, no.~04, 2020, pp.
  3609--3616.

\bibitem{wang2021dcn}
R.~Wang, R.~Shivanna, D.~Cheng, S.~Jain, D.~Lin, L.~Hong, and E.~Chi, ``Dcn v2:
  Improved deep \& cross network and practical lessons for web-scale learning
  to rank systems,'' in \emph{Proceedings of the Web Conference 2021}, 2021,
  pp. 1785--1797.

\bibitem{zhou2019deep}
G.~Zhou, N.~Mou, Y.~Fan, Q.~Pi, W.~Bian, C.~Zhou, X.~Zhu, and K.~Gai, ``Deep
  interest evolution network for click-through rate prediction,'' in
  \emph{Proceedings of the AAAI conference on artificial intelligence},
  vol.~33, no.~01, 2019, pp. 5941--5948.

\bibitem{li2019multi}
C.~Li, Z.~Liu, M.~Wu, Y.~Xu, H.~Zhao, P.~Huang, G.~Kang, Q.~Chen, W.~Li, and
  D.~L. Lee, ``Multi-interest network with dynamic routing for recommendation
  at tmall,'' in \emph{Proceedings of the 28th ACM international conference on
  information and knowledge management}, 2019, pp. 2615--2623.

\bibitem{pi2019practice}
Q.~Pi, W.~Bian, G.~Zhou, X.~Zhu, and K.~Gai, ``Practice on long sequential user
  behavior modeling for click-through rate prediction,'' in \emph{Proceedings
  of the 25th ACM SIGKDD International Conference on Knowledge Discovery \&
  Data Mining}, 2019, pp. 2671--2679.

\bibitem{pi2020search}
Q.~Pi, G.~Zhou, Y.~Zhang, Z.~Wang, L.~Ren, Y.~Fan, X.~Zhu, and K.~Gai,
  ``Search-based user interest modeling with lifelong sequential behavior data
  for click-through rate prediction,'' in \emph{Proceedings of the 29th ACM
  International Conference on Information \& Knowledge Management}, 2020, pp.
  2685--2692.

\bibitem{man2017cross}
T.~Man, H.~Shen, X.~Jin, and X.~Cheng, ``Cross-domain recommendation: An
  embedding and mapping approach.'' in \emph{IJCAI}, vol.~17, 2017, pp.
  2464--2470.

\bibitem{zhu2019dtcdr}
F.~Zhu, C.~Chen, Y.~Wang, G.~Liu, and X.~Zheng, ``Dtcdr: A framework for
  dual-target cross-domain recommendation,'' in \emph{Proceedings of the 28th
  ACM International Conference on Information and Knowledge Management}, 2019,
  pp. 1533--1542.

\bibitem{qiu2019cross}
M.~Qiu, B.~Wang, C.~Chen, X.~Zeng, J.~Huang, D.~Cai, J.~Zhou, and F.~S. Bao,
  ``Cross-domain attention network with wasserstein regularizers for e-commerce
  search,'' in \emph{Proceedings of the 28th ACM International Conference on
  Information and Knowledge Management}, 2019, pp. 2509--2515.

\bibitem{zhu2020deep}
F.~Zhu, Y.~Wang, C.~Chen, G.~Liu, M.~Orgun, and J.~Wu, ``A deep framework for
  cross-domain and cross-system recommendations,'' \emph{arXiv preprint
  arXiv:2009.06215}, 2020.

\bibitem{zhu2021unified}
F.~Zhu, Y.~Wang, J.~Zhou, C.~Chen, L.~Li, and G.~Liu, ``A unified framework for
  cross-domain and cross-system recommendations,'' \emph{IEEE Transactions on
  Knowledge and Data Engineering}, 2021.

\bibitem{yuan2019darec}
F.~Yuan, L.~Yao, and B.~Benatallah, ``Darec: Deep domain adaptation for
  cross-domain recommendation via transferring rating patterns,'' \emph{arXiv
  preprint arXiv:1905.10760}, 2019.

\bibitem{hu2018conet}
G.~Hu, Y.~Zhang, and Q.~Yang, ``Conet: Collaborative cross networks for
  cross-domain recommendation,'' in \emph{Proceedings of the 27th ACM
  international conference on information and knowledge management}, 2018, pp.
  667--676.

\bibitem{li2020ddtcdr}
P.~Li and A.~Tuzhilin, ``Ddtcdr: Deep dual transfer cross domain
  recommendation,'' in \emph{Proceedings of the 13th International Conference
  on Web Search and Data Mining}, 2020, pp. 331--339.

\bibitem{ouyang2020minet}
W.~Ouyang, X.~Zhang, L.~Zhao, J.~Luo, Y.~Zhang, H.~Zou, Z.~Liu, and Y.~Du,
  ``Minet: Mixed interest network for cross-domain click-through rate
  prediction,'' in \emph{Proceedings of the 29th ACM International Conference
  on Information \& Knowledge Management}, 2020, pp. 2669--2676.

\bibitem{li2021dual}
P.~Li, Z.~Jiang, M.~Que, Y.~Hu, and A.~Tuzhilin, ``Dual attentive sequential
  learning for cross-domain click-through rate prediction,'' \emph{arXiv
  preprint arXiv:2106.02768}, 2021.

\bibitem{zhu2022personalized}
Y.~Zhu, Z.~Tang, Y.~Liu, F.~Zhuang, R.~Xie, X.~Zhang, L.~Lin, and Q.~He,
  ``Personalized transfer of user preferences for cross-domain
  recommendation,'' in \emph{Proceedings of the Fifteenth ACM International
  Conference on Web Search and Data Mining}, 2022, pp. 1507--1515.

\bibitem{he2020dadnn}
J.~He, G.~Mei, F.~Xing, X.~Yang, Y.~Bao, and W.~Yan, ``Dadnn: Multi-scene ctr
  prediction via domain-aware deep neural network,'' \emph{arXiv preprint
  arXiv:2011.11938}, 2020.

\bibitem{sheng2021one}
X.-R. Sheng, L.~Zhao, G.~Zhou, X.~Ding, B.~Dai, Q.~Luo, S.~Yang, J.~Lv,
  C.~Zhang, H.~Deng \emph{et~al.}, ``One model to serve all: Star topology
  adaptive recommender for multi-domain ctr prediction,'' in \emph{Proceedings
  of the 30th ACM International Conference on Information \& Knowledge
  Management}, 2021, pp. 4104--4113.

\end{thebibliography}

% \vspace{12pt}
% \color{red}
% IEEE conference templates contain guidance text for composing and formatting conference papers. Please ensure that all template text is removed from your conference paper prior to submission to the conference. Failure to remove the template text from your paper may result in your paper not being published.

\end{document}